\documentclass[conference]{IEEEtran}
\IEEEoverridecommandlockouts
\usepackage{cite}
\usepackage{amsmath,amssymb,amsfonts}
\usepackage{algorithmic}
\usepackage{graphicx}
\usepackage{subcaption}
\usepackage{textcomp}
\usepackage{xcolor}
\usepackage{siunitx}
\usepackage{upgreek}
\usepackage{soul}
\usepackage{url}
\usepackage{dblfloatfix}
\usepackage{hyperref}
\def\BibTeX{{\rm B\kern-.05em{\sc i\kern-.025em b}\kern-.08em
    T\kern-.1667em\lower.7ex\hbox{E}\kern-.125emX}}
\begin{document}

\title{Improving Playtesting Coverage via Curiosity Driven Reinforcement Learning Agents}

\author{\IEEEauthorblockN{Camilo Gordillo, Joakim Bergdahl, Konrad Tollmar, Linus Gisslén}
\IEEEauthorblockA{
\textit{SEED - Electronic Arts (EA)}, Stockholm, Sweden \\
cgordillo, jbergdahl, ktollmar, lgisslen@ea.com}
}
\maketitle

\begin{abstract}
As modern games continue growing both in size and complexity, it has become more challenging to ensure that all the relevant content is tested and that any potential issue is properly identified and fixed. Attempting to maximize testing coverage using only human participants, however, results in a tedious and hard to orchestrate process which normally slows down the development cycle. Complementing playtesting via autonomous agents has shown great promise accelerating and simplifying this process. This paper addresses the problem of automatically exploring and testing a given scenario using reinforcement learning agents trained to maximize game state coverage. Each of these agents is rewarded based on the novelty of its actions, thus encouraging a curious and exploratory behaviour on a complex 3D scenario where previously proposed exploration techniques perform poorly. The curious agents are able to learn the complex navigation mechanics required to reach the different areas around the map, thus providing the necessary data to identify potential issues. Moreover, the paper also explores different visualization strategies and evaluates how to make better use of the collected data to drive design decisions and to recognize possible problems and oversights.

\end{abstract}


\section{Introduction}
Playtesting modern video games using human participants alone has become unfeasible due to the sheer scale of the projects. As games grow in size and complexity, maximizing coverage and ensuring sufficient exploration becomes a tedious, repetitive, and labor-intensive task. By contrast, automated approaches relying on AI-based agents have the potential to be parallelized and accelerated to provide results in a short period of time \cite{Unity}, thus complementing the regular testing pipelines.  

We tackle the problem of automatically exploring a given scenario with the purpose of identifying any potential issues which could, otherwise, be potentially overlooked by human testers. Human participants, we argue, should focus on testing and experiencing the key mechanics of the game without the burden of identifying, documenting and reporting general glitches. 

Our approach focuses on the use of reinforcement learning (RL) agents to maximize testing coverage. These types of agents have shown very important and appealing advantages over classical techniques when applied to game testing \cite{bergdahl} and may play an important role when working with complex 3D scenarios like the one presented in this paper. We make use of \textit{curiosity} as the motivation factor encouraging a set of RL agents to improve exploration and to seek novel interactions. It is therefore important to note that our intent is not to optimize a behaviour policy or any specific game score, but to make sure that proper and sufficient data can be collected while training such agents. Access to these kinds of data would enable a wide variety of applications such as automatically mapping reachable/unreachable areas in the scenario, identifying unintended mechanics, visualizing changes in response to design choices, to name a few. With the proper tools and metrics, moreover, other important issues like crash-inducing bugs and frame rate drops could also be triggered and recognized while training.

Once the game has been sufficiently explored and data has been collected, proper metrics and visualizations are required to make sense of the events recorded while training. Previous approaches have proposed different visualizations to derive insights about level design from playtesting data \cite{Agarwal20}\cite{Stahlke20}, and we use some of these ideas as reference to introduce metrics and analytics allowing us to validate our results and to identify different problems around the environment.

\section{Related Work}

To date, a couple of studies have investigated the use of automatic exploration techniques to maximize game state coverage. \textit{Walk Monster} \cite{Muratori18} is an automated reachability testing tool implemented while developing \textit{The Witness} (released in 2006 as a puzzle game). The purpose of this tool was to validate the traversability of the map and to identify any potential issues: players getting stranded by reaching areas they were not supposed to get into, getting stuck inside geometries, etc. The proposed algorithm managed to achieve impressive results despite employing fairly simple exploration heuristics. Nevertheless, it is important to note the simplicity and low dimensionality of the game itself (a two-dimensional space in practice). Similarly, several exploration strategies have been evaluated with the aim of producing a semantic map of reachable states in several commercial games (from Atari 2600 to Nintendo 64) \cite{ZhanAS19}. Even though their results are comparable to human gameplay, their exploration strategies rely heavily on random actions. This, we argue, would not be applicable in more complex scenarios like the one presented in this paper. The \textit{Wuji} \cite{Wuji19} framework, on the other hand, employs a RL policy similar to ours together with evolutionary multi-objective optimization to encourage exploration and high game state coverage in two commercial combat games. Contrary to our approach, however, the authors do not evaluate the use of data and visualizations to allow for the identification of bugs and oversights.

More recent approaches have been designed to take advantage of human demonstrations. When data from human participants is available, the \textit{Reveal-More} algorithm \cite{ChangAS19} can be used to amplify coverage by focusing on exploring around those trajectories. This idea is indeed very interesting and orthogonal to our current approach. We could, for example, make use of imitation learning techniques to encourage and bias exploration around those human generated trajectories. 

Another similar research direction focuses on the use of \textit{procedural personas} to mimic how different player archetypes would interact with a given scenario. The \textit{PathOS} framework \cite{Stahlke20}, for example, is a very comprehensive tool for simulating testing sessions with artificial agents. Each of these agents is modeled to represent one particular player archetype by using classical scripted AI. Scripting each of these behaviours, however, can prove to be quite challenging as the complexity of the game increases. Other approaches, in contrast, have tried to automate the generation of such behaviours. The authors of \cite{holmgaard2018automated}, for example, propose the use of Monte Carlo tree search and evolutionary algorithms to generate utility functions leading to different behaviours in 2D dungeon levels. How this approach would scale to games of higher complexity, however, remains an open question.  

Regardless of what kinds of automated strategies are employed proper visualizations and metrics are key to make sense of the collected data. In \cite{Agarwal20} the authors propose a set of visualizations to analyze level design in 2D side-scrolling games. In a similar fashion, the authors of \cite{Chang_Smith_2020} introduce \textit{Differentia}, a set of visualizations to evaluate incremental game design changes. Although our approach draws inspiration from all of these techniques, it is important to remark that 2D visualizations are unlikely to be enough when testing complex 3D scenarios. The \textit{PathOS} framework \cite{Stahlke20}, on the other hand, is perhaps one of the most similar approaches in terms of visualizations and metrics allowing the user to visualize the outcome of the simulation directly in the game engine.  

Meanwhile, \textit{intrinsic motivation} is a highly studied topic within the reinforcement learning community aiming to encourage agents to explore and to play in the absence of an \textit{extrinsic} reward. One of such motivations, \textit{curiosity}, was originally proposed by \cite{schmidhuber1991possibility} as a way of rewarding the agents for exploring previously unseen game states and for improving their knowledge about the world. In \cite{pathak2017curiosity}, curiosity is used as a mechanism for pushing the agents to explore complex environments more efficiently while learning skills which may become useful later in their lifetime. A detailed survey about the use of intrinsic motivation in reinforcement learning is presented in \cite{Aubret19}.
 
\section{Implementation}
\label{sec:implementation}

Our approach relies on a set of RL agents continuously interacting with the game and encouraged to maximize coverage. In contrast to the \textit{PathOS} framework \cite{Stahlke20} (see above), we employ \textit{curiosity} as the sole motivation profile driving each of the agents. The following sections describe the scenario developed for our experiments, the RL setup and training algorithm, and the tools which were developed to collect relevant data and generate visualizations directly in the game engine. 

\subsection{Environment}

\begin{figure}[!t]
    \centering
    \includegraphics[width=0.45 \textwidth]{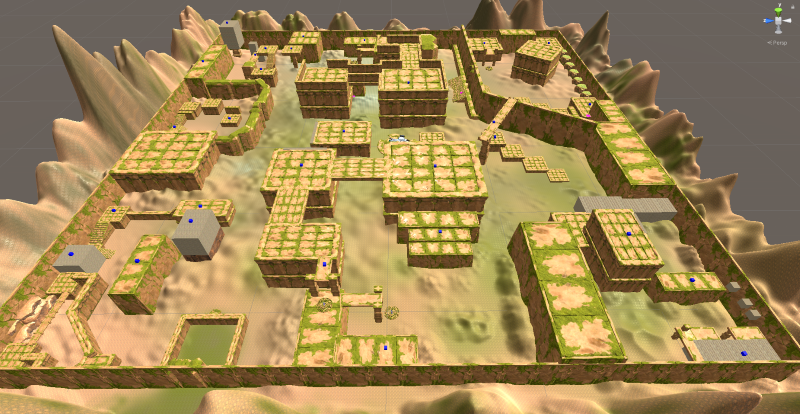}
    \caption{Evaluation map: \SI{500}{\metre} x \SI{500}{\metre} x \SI{50}{\metre}. The environment contains complex navigation challenges composed of multiple sequential jumps, climbable objects and elevators.}
    \label{fig:LargeScenario}
\end{figure}

We evaluate our approach on a relatively large (\SI{500}{\metre} x \SI{500}{\metre} x \SI{50}{\metre}) map designed for the purpose of creating an elaborate navigation landscape. We created a scenario with complex navigation mechanics (e.g. jumps, climbable walls and elevators) in a 3D space as shown in Fig. \ref{fig:LargeScenario}. Moreover, and similar to what it is normally seen in modern video games, we have designed the scenario so that complex navigation strategies are required to fully explore the different areas around the map. More details about the environment, navigation mechanics and final results can be found in the accompanying video\footnote{\url{https://www.youtube.com/watch?v=cfm3R94FB_4}}.

The character in our environment (see Fig. \ref{fig:Character}) is \SI{1.7}{\metre} tall and has a total of 3 continuous navigation actions: forward/backward, left/right turn, left/right strafe, plus a discrete action for jumping. The character also has the ability to climb on special surfaces located around the map.

Because the purpose of our approach is to test the game and identify any potential issues, we introduced a set of known bugs into the map. These bugs include missing collision boxes, design oversights, places where players may get stuck, etc.

It is important to remark that, contrary to similar approaches like the one presented in \cite{Stahlke20}, we make no use of navigation meshes within our environment. As discussed in \cite{bergdahl}, navigation meshes are often not designed to resemble the freedom of movement that human players will have. Any agent constrained by these meshes will, most likely, fail at exploring the environment to the same degree a human would, and will therefore miss those bugs frequently found by human players. In the next section we discuss how an RL agent can be used to control navigation and improve exploration.

\subsection{Reinforcement learning setup}

\begin{figure}[!b]
    \centering
    \includegraphics[trim=500 100 500 100, clip, width=0.30 \textwidth]{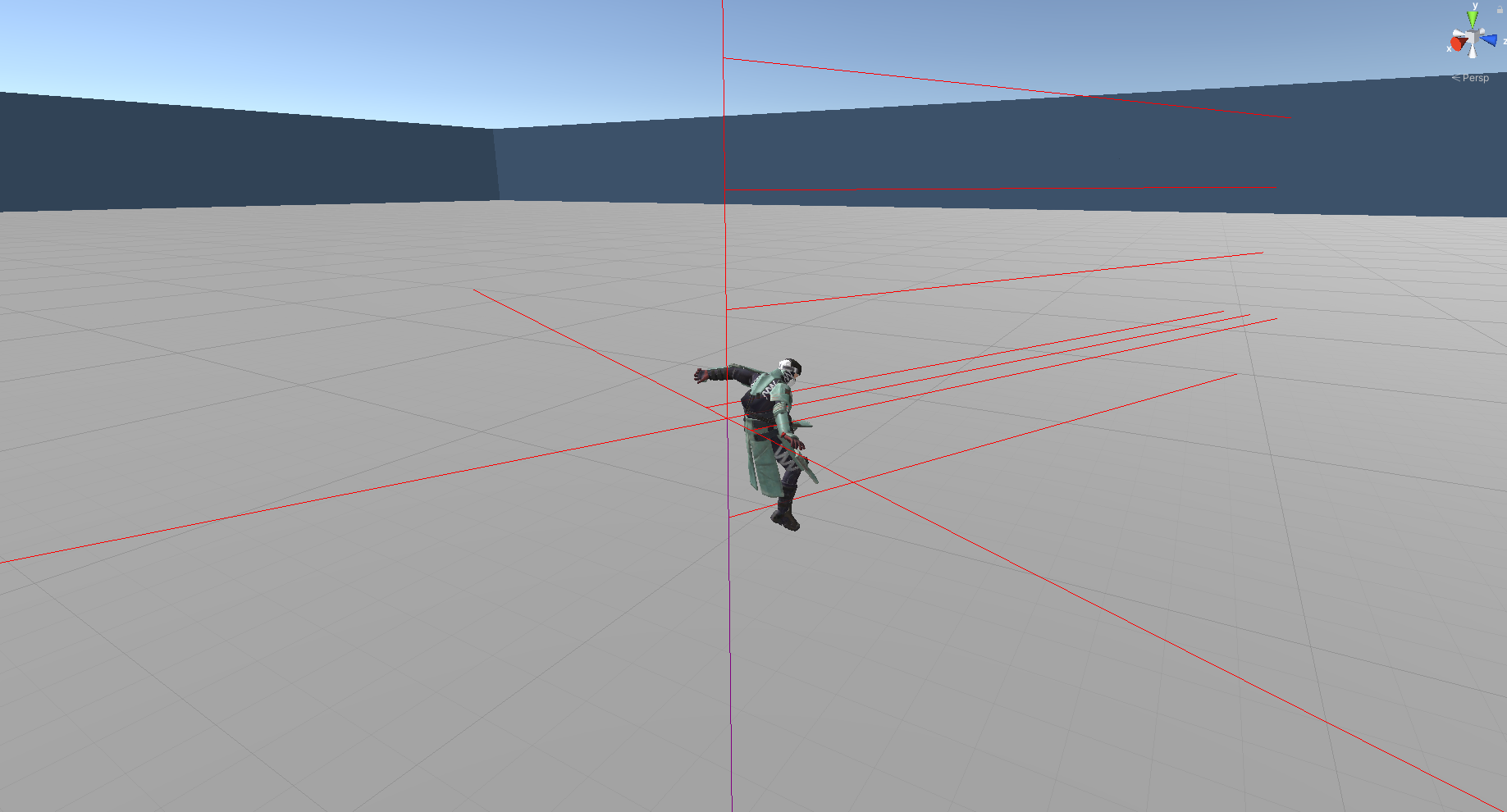}
    \caption{A representation of our character and the ray casts composing the vision array.}
    \label{fig:Character}
\end{figure}

We make use of proximal policy optimization (PPO) \cite{schulman2017proximal} as our RL training algorithm. PPO is a robust and well established baseline within the RL community when working with continuous action spaces. We evaluate and compare the performance of the agents when providing two different types of observations. The first one consists of an aggregate vector of 37 values: agent position ($\mathbb{R}^{3}$), agent velocity ($\mathbb{R}^{3}$), agent world rotation ($\mathbb{R}^{4}$), is climbing ($\mathbb{B}$), is in contact with ground ($\mathbb{B}$), jump cool-down time ($\mathbb{R}$), and a vision array ($\mathbb{R}^{24}$). The vision array consists of 12 ray casts projected in various directions around the agent (see Fig. \ref{fig:Character}). Each of these rays provides two values: a collision distance and a semantic meaning depending on the type of object it collides with. All values are normalized to be kept between $[-1, 1]$. We also explore a second configuration by providing the agents with an additional first person view of the environment and we compare both models in Section \ref{sec:coverage}.

\begin{table}[!t]
\centering
\begin{tabular}{ll}
\hline
\multicolumn{2}{c}{Global Hyperparameters}   \\ 
\hline
\hline
{\it Name}        & {\it Value}    \\ 
\hline
Learning rate ($\upalpha$)  & 1e-4           \\
Discount ($\upgamma$)       & 0.98           \\
PPO-Clip                    & 0.2            \\
Entropy coefficient         & 1e-2           \\
GAE coefficient ($\uplambda$) & 0.95         \\
Fully connected layers      & [1024, 512, 256]      \\
LSTM layer                  & 256      \\
\hline
\hline
\multicolumn{2}{c}{Visual Encoder }   \\ 
\hline
Image size & [84,84,3]          \\
Kernel size & [5,3,3,3]         \\
Padding     & [1,1,1,1]         \\
Strides     & [2,2,2,1]         \\
Channels     & [32,32,64,64]    \\
\hline
\end{tabular}
\caption{Algorithm's training hyperparameters and model architecture.}
\label{tab:algoparams}
\end{table}

The reward given to the agents is a function of \textit{novelty} and it is described, together with the reset logic, in the following sections. The algorithm's hyperparameters and the model's architecture are presented in Table \ref{tab:algoparams}.

\subsection{Optimizing coverage through count-based exploration}
\label{sec:reward}

One of the great advantages of automated testing is the ability to parallelize and scale to a degree which is unfeasible to reach with human participants alone. The results described in this paper were collected while simulating and training 320 agents distributed across multiple machines. Our distributed setup allows us to train a single and centralized model using data collected from multiple environment instances. Instantiating a single training server also allows us to easily process, analyze, and store all the data collected by the agents in a single place. 

The reward given to the agents is computed following the idea of count-based exploration \cite{Bellemare16} and becomes, therefore, inversely proportional to how frequent a given game state has been visited. We define these states as the 3D position of the agent at a given point in time. Keeping track of these visit counters on a continuous space, however, quickly becomes intractable. To solve this problem, we discretize the space by means of a threshold $\uptau$. An agent is only considered to have entered a new state once its distance to any previously visited state is larger than $\uptau$. 

Using a small value for $\uptau$ increases the number of points we will need to keep track of and may therefore hinder performance when working with large maps. A high value, on the contrary, results in a very sparse reward signal for our agents which significantly increases the difficulty of the task. The value of $\uptau = \SI{5}{\metre}$ was empirically found after a couple of experiments and has proven to be a sensible choice not only on the scenario presented here, but also in other maps not shown in this paper.

When a new observation is received, the first step to compute a reward is to extract the position $p$ of the agent and compare it to all the previously visited states. This \textit{buffer} is initially empty and gets populated as exploration takes place. We also keep track of a visit counter $N_{i}$ for each point $i$ in the current buffer. If the minimum distance between the current position $p$ and the points in the buffer is larger than $\uptau$, then point $p$ is added to the buffer and its visit counter is set to 1. If, on the contrary, the minimum distance is smaller than $\uptau$, we identify the point $i$ within the buffer closest to $p$ and increment $N_{i}$ by 1. Having done this, the reward for reaching point $i$ is computed using Equation \ref{eq:reward}, where $R_{max}$ is set to 0.5 and defines the reward for exploring a new point, and $max\_counter$ is set to 500, thus annealing the reward down to zero as a given point gets more frequent visits.

\begin{equation}
R_t = R_{max} * \left[1 - \frac{N_{i}}{max\_counter}\right]
\label{eq:reward}
\end{equation}

\subsection{Reset logic}

Each training episode is simulated for 3000 steps (equivalent to 1 minute of game play) and the agents are respawned once time is up. An initial spawning location was defined for this scenario and is located near the middle of the map at ground level. As training goes on spawning locations are sampled from the current buffer using the inverse of the corresponding visit counters as sample weights. This logic prevents biasing the exploration by respawning the agents in previously unexplored positions while giving priority to less frequently visited locations.  

To prevent agents from spawning in mid-air, we take advantage of one of the values available in the observation vector: $is\_in\_contact\_with\_ground$. When storing a new point in the buffer we keep track of whether or not the character was stepping on something when at that location. Then, when sampling a new spawn position, we can just consider the points in the buffer for which this condition is true. 

This way of respawning the agents at previously visited states strongly resembles algorithms such as Rapidly-Exploring Random Trees (RRT) \cite{LaValle1998}. In contrast to the exploration strategies proposed in \cite{ZhanAS19}, however, we take advantage of the complex navigation strategies developed by our agents to explore around those spawning locations. In Section \ref{sec:results} we compare the performance of our approach to the one of a random policy very similar to the \textit{chaos monkey} strategy proposed in \cite{ZhanAS19}. This random policy employs the same respawning logic presented above which allows us to fairly compare both techniques.

\subsection{Collecting and visualizing data}

Different kinds of data are continuously processed and stored while the agents interact with the environment. Most of this data is handled by the centralized training process which receives all episodic information (i.e. observations, actions, rewards). Some of the data, however, is recorded directly by the environments based on possible events triggered by the agents which are harder to identify outside the engine. The nature of this data, and how it is used to identify and correct problems, is described together with the corresponding experiments in the next section. 

Other relevant metrics such as the number of visited states and values relevant to the RL algorithm are also continuously logged and visualized as training goes on. These logs are very useful to quickly judge and/or compare the performance of a set of experiments without the need of waiting until their completion. 

\section{Results}
\label{sec:results}

In this section we present results on the exploration performance of our agents and give a few examples on the type of analyses which can be conducted using the collected data. As described in the previous section, our simulation pipeline generates a set of files which can be loaded directly into the game engine allowing us to identify potential problems in the game.

\subsection{Exploration performance and map coverage}
\label{sec:coverage}

\begin{figure}[!t]
    \centering
    \includegraphics[width=0.45\textwidth]{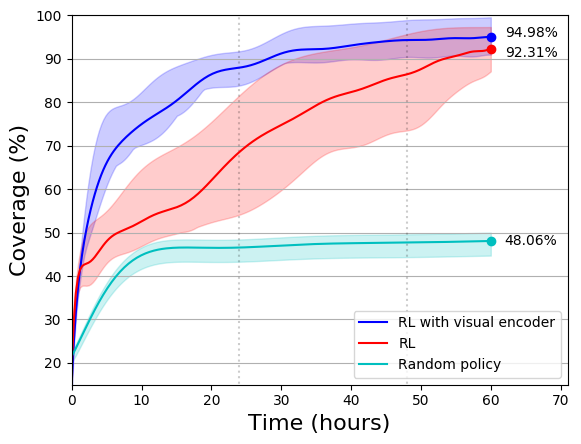}
    \caption{Map coverage as a function of simulation time. The maximum number of points which could be reached (equivalent to 100\%) was estimated to be 25K. The plot shows the mean and variance of the performance of different policies over 5 different runs.}
    \label{fig:Coverage}
\end{figure}

The first thing we would like to evaluate is the ability of our RL agents to navigate and explore the whole map. To do this, we make use of the \textit{buffer} of visited states introduced in Section \ref{sec:reward}. This set of visited 3D coordinates is stored and updated as training goes on and can be used as a metric for exploration and coverage. Fig. \ref{fig:Coverage} shows the percentage of the map covered by our agents when compared to a random policy. As expected, the random based exploration technique did not cover the whole map due to its complexity and was only able to reach easily accessible areas. Moreover, complementing the observation space of our RL agents with a camera image (first person view) improves the results by allowing the model to better understand its surroundings and by decreasing the uncertainty introduced by the discrete set of ray casts. It currently takes around 24 hours to explore 90\% of the map but, as discussed in Section \ref{sec:conclusion}, we believe that coming up with better and more efficient ways for encoding the environment may boost the performance of the agents and speed up exploration. 

\begin{figure}[!t]
    \centering
    \begin{subfigure}{0.45\textwidth}
        \centering
        \includegraphics[width=\textwidth]{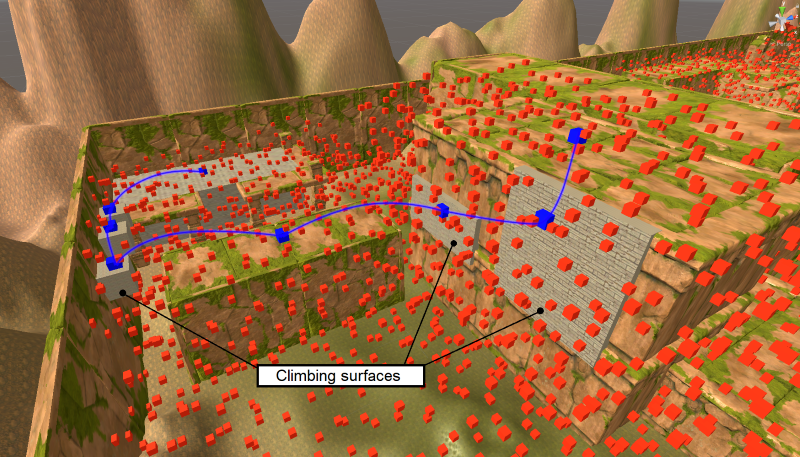}
        \caption{Challenge A: The agents were able to reach the top at the right of the figure by jumping and climbing over a series of obstacles.}
    \end{subfigure}
    \bigskip
    \begin{subfigure}{0.45\textwidth}
        \centering
        \includegraphics[width=\textwidth]{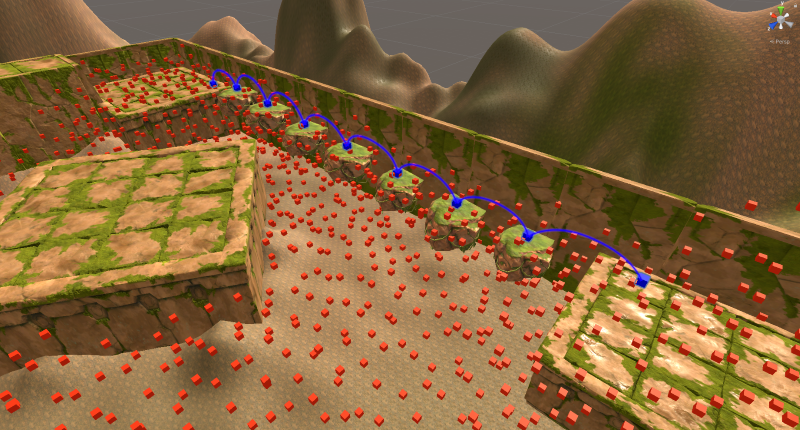}
        \caption{Challenge B: The only way of reaching the top of the block to the left is by sequentially jumping over the rocks.}
    \end{subfigure}
    \caption{Two of the navigation challenges spread across the map and the solution found by our training RL agents. The red cubes represent the points in our buffer of visited states. The blue trajectories showcase the paths that a player would have to follow to fully explore these areas.}
    \label{fig:nav_challenges}
\end{figure}

\begin{figure}[!t]
    \centering
    \begin{subfigure}{0.45\textwidth}
        \centering
        \includegraphics[width=1.0 \textwidth]{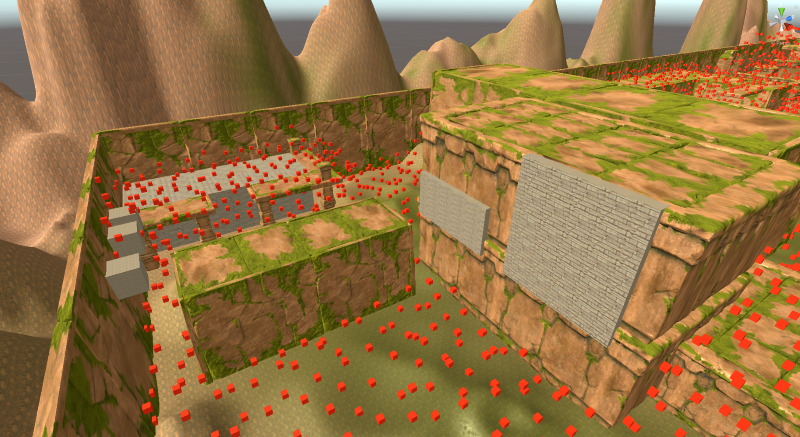}
        \caption{Challenge A: A random policy was, as expected, unable to solve complex navigation tasks even when simulated for longer periods of time.}
    \end{subfigure}
    \bigskip
    \begin{subfigure}{0.45\textwidth}
        \centering
        \includegraphics[width=1.0 \textwidth]{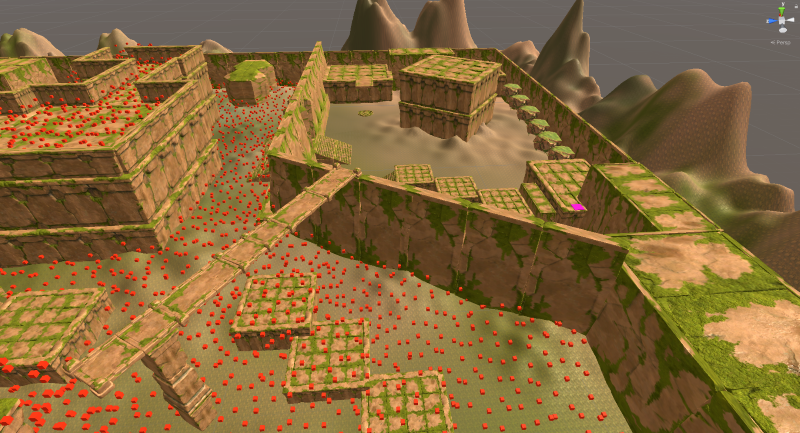}
        \caption{Challenge B: The only entry point to the top right area in this challenge is via the bridge coming from the lower-left. Random agents were unable to find this path.}
    \end{subfigure}
    \caption{Exploration performance of a random exploration strategy when faced with complex navigation challenges.}
    \label{fig:nav_challenges_random}
\end{figure}

As shown in Figs. \ref{fig:nav_challenges} and \ref{fig:nav_challenges_random}, all the data collected by the agents (e.g. the buffer of visited states) can also be loaded and displayed on top of the map. These visualizations allow the designers to verify whether or not different areas across the map are reachable and also allow us to compare the exploration performance of different navigation strategies. Both figures showcase a couple of relatively complex navigation challenges spread across the map and the extend of the exploration coverage achieved by our RL agents when compared to a simple random exploration strategy. 

Another set of interesting findings relates to the agents reaching areas which should have been inaccessible for the player. These areas could be identified either by visually inspecting the distribution of the collected points or, as shown in the next section, by defining exploration boundaries.

\subsection{Exploration boundaries and regions of interest}

Our method allows the designer to specify both an exploration boundary (EB) and regions of interest (ROIs) across the map prior to training. The EB defines the section of the map which should be explored and the episode terminates whenever the agent exits that boundary. The ROIs, on the other hand, are optional regions inside the EB and serve as a reference for data collection.

Although we would like to record and store the specific trajectories followed by the agents, doing so would quickly become too expensive and intractable during long simulations. The definition of the EB and the ROIs, however, allows us to focus on those trajectories which are likely to be useful for testing the game. Our technique keeps track of the episodic trajectory for each agent but only records them if a couple of conditions are met: first, the trajectory needs to cross over the boundary defining either the EB or a ROI; second, the point at which the agent crosses that boundary needs to be significantly different to the one of any previously recorded trajectories. 

\begin{figure}[!t]
    \centering
    \includegraphics[width=0.45\textwidth]{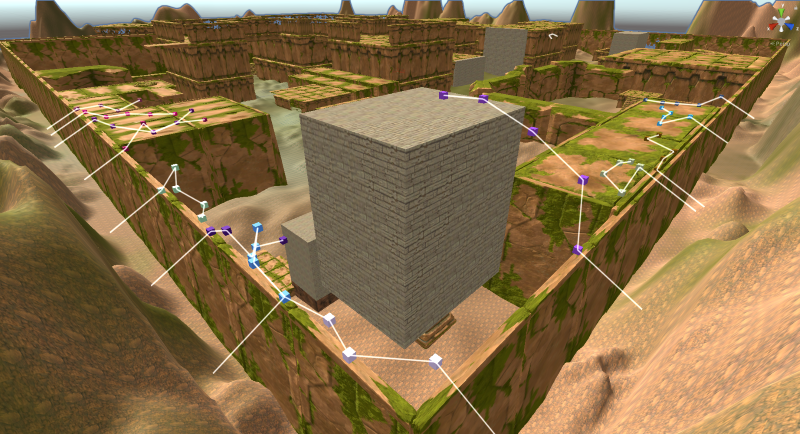}
    \caption{Visualization of trajectories leaving the exploration boundary. The agents have found multiple ways of exploiting design oversights to reach into areas which should be inaccessible. These trajectories were recorded during training and could be used to redesign the map and fix these issues.}
    \label{fig:LeaveEB}
\end{figure}

\begin{figure}[!t]
    \centering
    \includegraphics[width=0.45\textwidth]{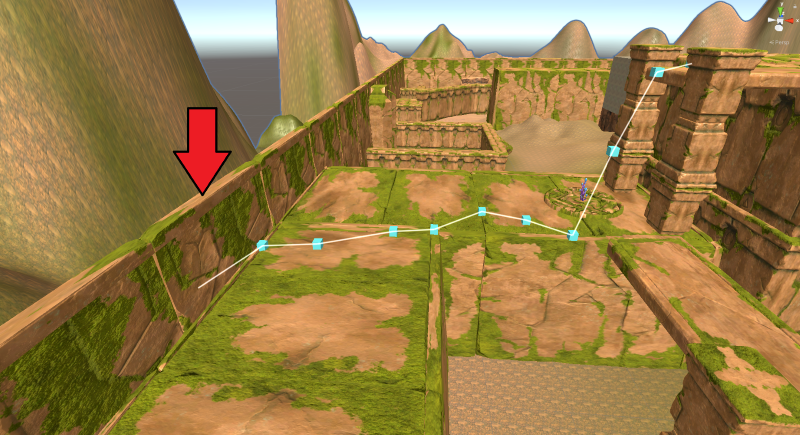}
    \caption{A trajectory leaving the game area due to a missing collision box. The collision box of a small segment in the wall (red arrow) was intentionally removed to validate the use and application of the collected data and the proposed visualizations.}
    \label{fig:LeaveEB2}
\end{figure}

\begin{figure}[!t]
    \centering
    \includegraphics[width=0.45\textwidth]{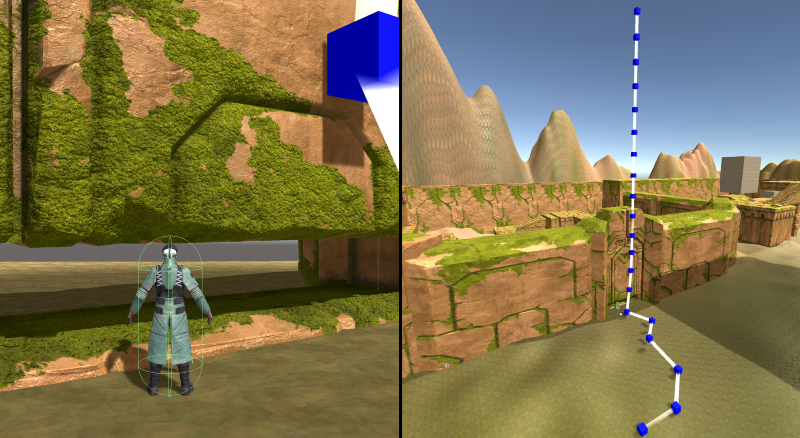}
    \caption{The agents were getting stuck in the gap between these objects (left) which was causing the physics engine to eventually throw them upwards into the sky. A trajectory leaving the exploration boundary (right) was recorded by our system and allowed us to fix the problem by closing up the gap.}
    \label{fig:fly_glitch}
\end{figure}

Fig. \ref{fig:LeaveEB} shows examples of trajectories leaving the EB when that boundary is defined at the walls surrounding the scenario in Fig. \ref{fig:LargeScenario}. Our technique allows the user to display these trajectories directly in the game engine and, in this case, would reveal design oversights around the map allowing the player to leave the game area. Figs. \ref{fig:LeaveEB2} and \ref{fig:fly_glitch} show additional examples of the kinds of issues which could be identified using these visualizations. Fig. \ref{fig:LeaveEB2} shows a trajectory leaving the scenario due to a collision box missing in one of the wall segments. This particular oversight was intentionally introduced in the map to validate the usefulness of the collected data. Interestingly, not all of the problems we were able to identify were intentionally added to the game. Fig. \ref{fig:fly_glitch}, for example, shows a trajectory recorded during some of our first design iterations. In this case, the agent was getting stuck in between two objects and the physics engine would eventually throw it upwards forcing the character to leave the EB. This shows the use these visualizations could have for identifying problems early on in the design process.

\begin{figure}[!t]
    \centering
    \includegraphics[width=0.45\textwidth]{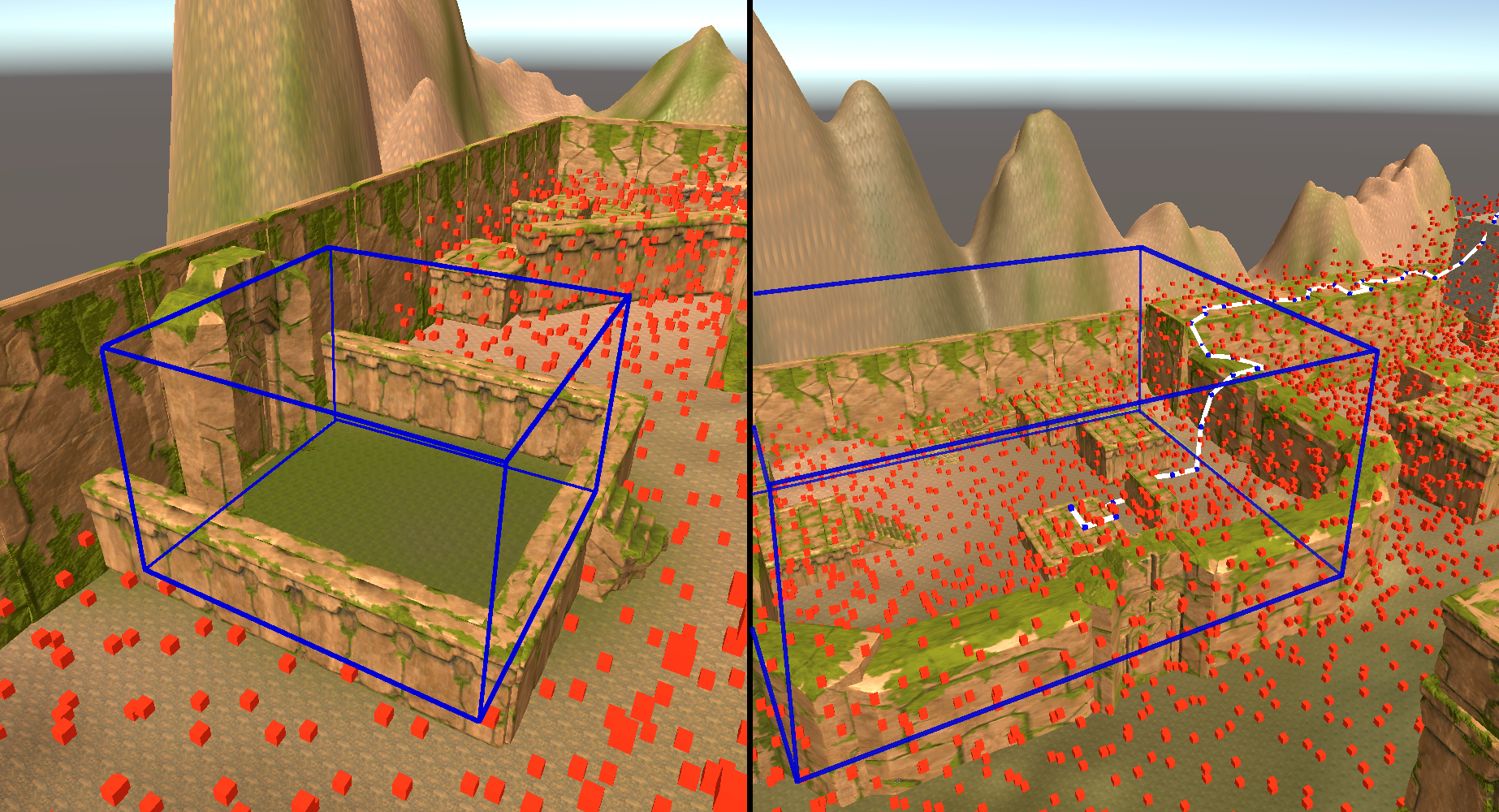}
    \caption{Two regions of interest defined at areas which should have been unreachable for the player. The buffer of visited states (red cubes) allows us to see that the region to the left remained unexplored while the region to the right was reachable somehow. The trajectories recorded while training, however, allow us to easily visualize how the agents did manage to break into that region and could help the designers to correct any oversight.}
    \label{fig:ROIs}
\end{figure}

The ROIs, on the other hand, allow the user to validate the reachability and access to particular areas in the map. Fig. \ref{fig:ROIs} shows two such regions which were intended to be unreachable for the player. The agents were indeed unable to reach the first region and therefore no trajectories were recorded. The second region, however, ended up having one access point which could be identified using the collected data.

\subsection{Connectivity graph}

\begin{figure}[!t]
    \centering
    \includegraphics[width=0.45\textwidth]{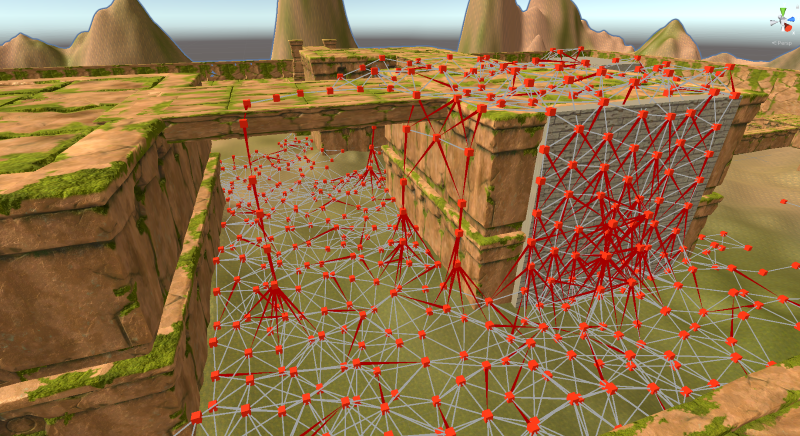}
    \caption{Visualizing the connectivity between a small subset of collected points. Bidirectional edges are represented in white while red edges point towards the target node. This connectivity graph broadly behaves like a classical navigation mesh and represents the exploration space that the players are able to traverse.}
    \label{fig:Connectivity}
\end{figure}

We can do more than just storing a point cloud of visited states. We can, for instance, configure our training server to build and store a graph structure representing the connectivity between those points. For this reason, the episodic trajectories collected to train our agents are also used to generate a directed graph as the one shown in Fig. \ref{fig:Connectivity}. Even though the accuracy of such a graph strongly depends on our discretization threshold $\uptau$ (see Section \ref{sec:reward}), we believe it has a couple of very promising applications as presented next.

\bigskip
\subsubsection{Navigating to custom points of interest}

\begin{figure}[!t]
    \centering
    \includegraphics[width=0.45\textwidth]{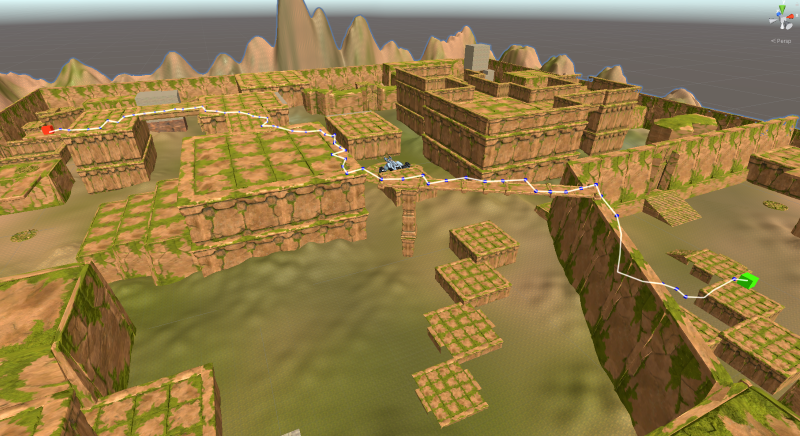}
    \caption{The connectivity graph presented in Fig. \ref{fig:Connectivity} allows us to generate trajectories between any given two points in the map. In this case, the red block to the left was set as the origin while the green block to the right was declared as the navigation target. The path which an agent could have followed is shown in white.}
    \label{fig:path_gui}
\end{figure}

We can make use of the connectivity graph and path planning algorithms to estimate navigation trajectories between any given two points. Our tool allows the user to define an initial and a target position in the map and it then generates a navigation trajectory between those two points based on the data collected from the agents. Fig. \ref{fig:path_gui} shows an example of such a trajectory. 

\begin{figure}[!t]
    \centering
    \begin{subfigure}{\linewidth}
        \centering
        \includegraphics[width=0.9 \textwidth]{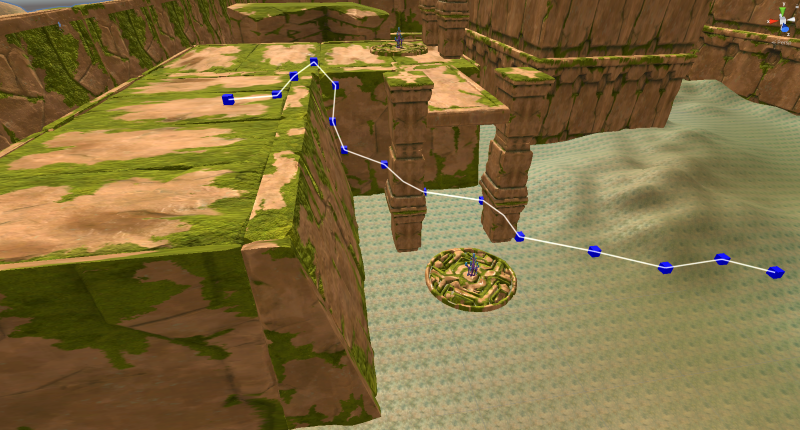}
        \caption{There seems to be a path leading to the top of this platform without the need of using the elevator. The problem is caused by the slope of the walls at that particular corner.}
        \label{fig:climb_glitch_a}
    \end{subfigure}
    \bigskip
    \begin{subfigure}{\linewidth}
        \centering
        \includegraphics[width=0.9 \textwidth]{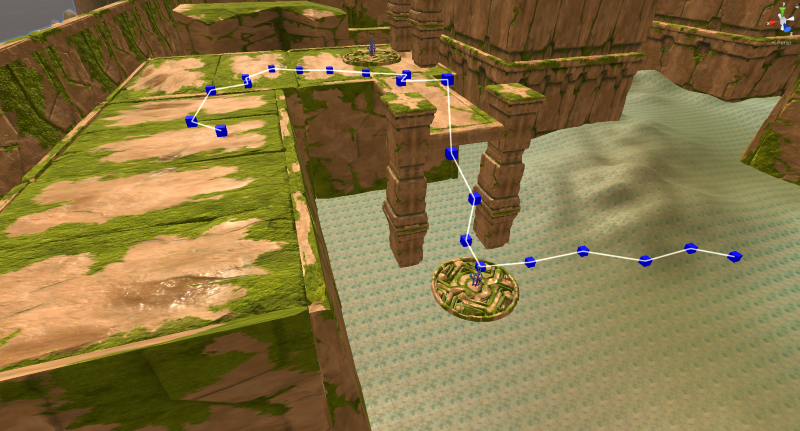}
        \caption{This is how the same path looks like once the slopes are adjusted to prevent the agents from climbing. Taking the elevator is now the only option to reach the top.}
        \label{fig:climb_glitch_b}
    \end{subfigure}
    \caption{The connectivity graph and the visualization of custom trajectories allowed us to identify a minor oversight resulting in agents being able to climb over this particular segment (more details in the accompanying video).}
    \label{fig:climb_glitch}
\end{figure}

We argue that a tool like this could be very useful for designers to comprehend how the agents are navigating the map and whether or not they have found potential exploits. Fig. \ref{fig:climb_glitch}, for example, shows one particular exploit which \textit{wasn't} intentionally introduced into the game and which was identified thanks to the connectivity graph. In Fig. \ref{fig:climb_glitch_a} the agents are able to climb over the wall without the need of the elevator. This seems to be happening due to the slope of the walls at that particular corner (you can see more details in the accompanying video). Fig. \ref{fig:climb_glitch_b}, in contrast, shows the trajectory followed by the agent once the previous issue was fixed. 

\subsubsection{Semantic connectivity maps}

\begin{figure}[!t]
    \centering
    \includegraphics[width=0.45\textwidth]{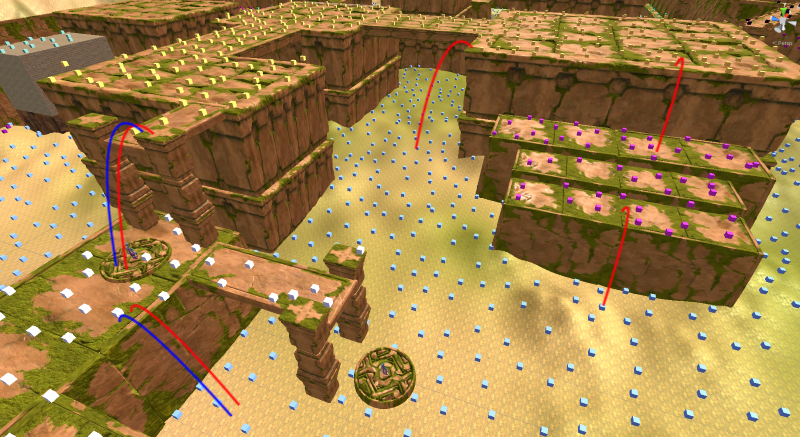}
    \caption{Semantic map generated automatically using the data collected from the agents. Regions in the map are identified and color-coded using unsupervised clustering algorithms and the connectivity graph is then used to visualize traversability between regions. Blue and red lines represent upwards and downwards trajectories respectively.}
    \label{fig:clustering}
\end{figure}

The connectivity graph can also be used to identify how different areas in the map are connected to each other. The specific regions in the map could be either manually defined by the user or, as in our next experiment, automatically extracted from the point cloud of visited states. Fig. \ref{fig:clustering} shows an example of such a mapping where the regions in the map were automatically extracted and color-coded using unsupervised clustering algorithms. Once the regions are identified, we can make use of the connectivity graph to analyze what kinds of connections exist between them. This semantic mapping could then be used to drive design decisions, to validate the mechanics and traversability of the map, and to recognize potential exploits (unexpected paths). 

\subsection{Termination states}

It is common for players to find themselves stuck in some particular part of the map due to issues in the environment or the location of the game assets which render the playing character immovable. Maximizing exploration coverage gives us the opportunity to automatically identify such locations during training. One approach is to keep track of a termination counter for each point in our buffer of visited states (i.e. how many times an episode ended with an agent in that location). Once training is over, we can proceed and analyze the distribution of terminal states. Any outlier in this distribution can be easily identified and it is likely to be caused by the agents getting stuck in that position. 

\begin{figure*}[!t]
    \centering
    \begin{subfigure}{0.33\linewidth}
        \centering
        \includegraphics[width=1.0 \textwidth]{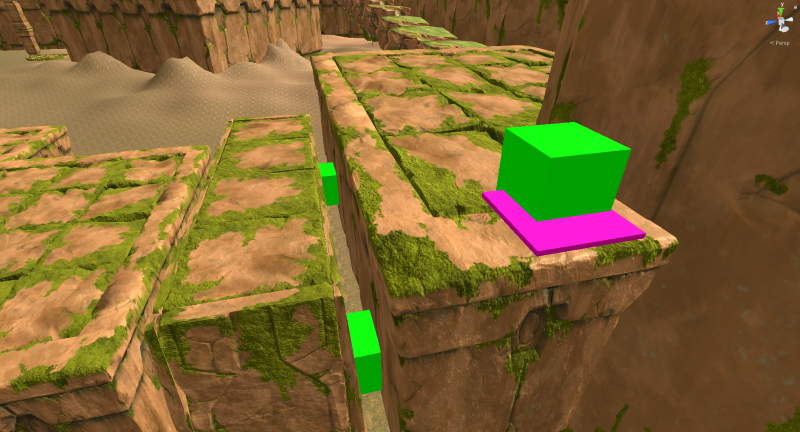}
        \caption{}
        \label{fig:stuck1}
    \end{subfigure}\hfill
    \begin{subfigure}{0.33\linewidth}
        \centering
        \includegraphics[width=1.0 \textwidth]{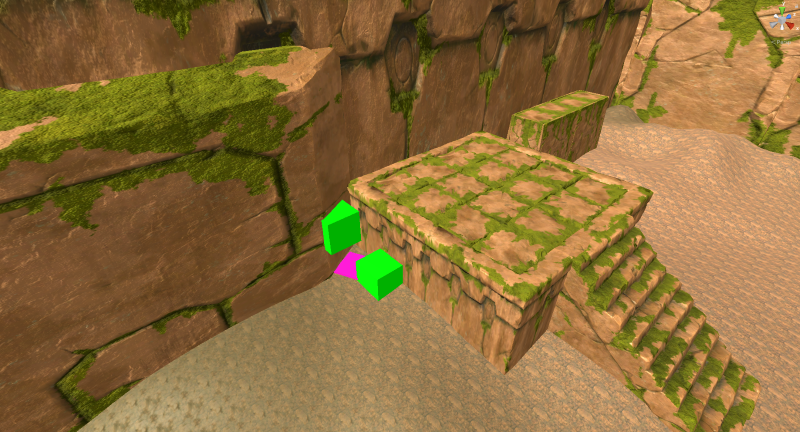}
        \caption{}
        \label{fig:stuck2}
    \end{subfigure}\hfill
    \begin{subfigure}{0.33\linewidth}
        \centering
        \includegraphics[width=1.0 \textwidth]{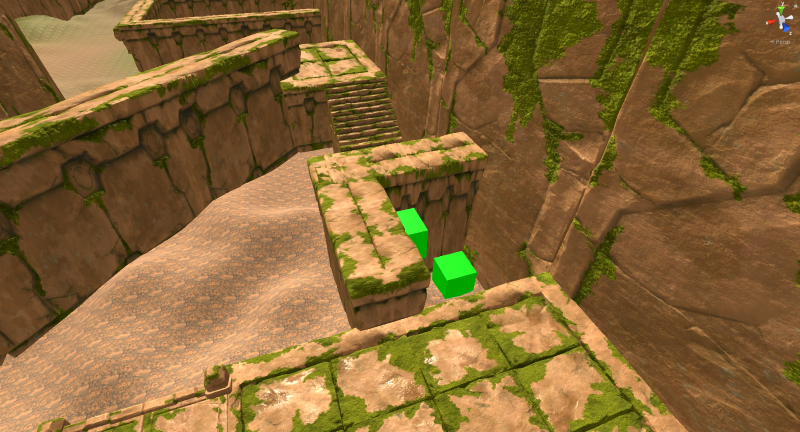}
        \caption{}
        \label{fig:stuck3}
    \end{subfigure}
    \caption{Visualizing areas where players could get stuck. The purple areas are intentionally introduced surfaces which freeze the player in place if they come in contact. The green blocks represent the outliers encountered in the distribution of terminal states and highlight the locations around the map which could be problematic. Figure (\subref{fig:stuck1}) shows two additional blocks between two platforms which were caused by the agents falling in the gap and getting stuck. This was a design oversight which was not intentionally introduced into the game and which was identified thanks to these visualizations. Figure (\subref{fig:stuck3}), on the contrary, shows a region in the map which was designed for the agents to get trapped if they fell into it.}
    \label{fig:stuck_figs}
\end{figure*}

We conducted experiments by introducing areas across the map where the agents could get stuck. Fig. \ref{fig:stuck_figs} shows some examples of such locations together with the outlier positions identified from the collected data. Due to the high coverage achieved by our agents, all the intentionally introduced issues could be identified, as well as one unintentional design oversights causing a similar problem (see Fig. \ref{fig:stuck1}). 

\section{Conclusions and future work}
\label{sec:conclusion}

In this paper we have shown a potential use case for RL agents trained to maximize testing coverage in complex 3D scenarios. We have reported on the use of \textit{curiosity} to encourage exploratory behaviour in our agents, thus allowing them to fully traverse the environment. We have shown that curiosity driven agents can be used for automating the collection of playtest data and performance metrics. 

The aim of our approach was to maximize testing coverage and to keep time consumption to a minimum by means of scaling and parallelizing data collection. We have provided examples on the type of data which can be collected, the kind of analysis which can be conducted, and the different sets of visualizations and metrics which can be used to facilitate the identification of frequent oversights, glitches and exploits.

A natural progression of this work is to further increase the complexity of the environment by introducing new mechanics, objectives and environmental hazards. This line of research is also strongly dependent on finding better and more efficient ways of encoding the environment. As discussed in Section \ref{sec:coverage}, the way the agents perceive the map both influences the complexity of the task and the cost of training.

Another promising research vector relates to the use of human demonstrations. One could, on one hand, explore a similar idea to the one presented in \cite{Chang_Smith_2020} and focus on exploring around predefined human-generated trajectories. This approach will provide designers with more control over the exploration space and will therefore speed up coverage over regions of high interest. On the other hand, human demonstrations could also be used, together with imitation learning, to provide the agents with some prior understanding about the mechanics of the game and with some basic navigation skills. This prior knowledge is then likely to speed up exploration and decrease the time it takes to collect relevant data.

\bibliography{refs}

\begin{thebibliography}{10}
\providecommand{\url}[1]{#1}
\csname url@samestyle\endcsname
\providecommand{\newblock}{\relax}
\providecommand{\bibinfo}[2]{#2}
\providecommand{\BIBentrySTDinterwordspacing}{\spaceskip=0pt\relax}
\providecommand{\BIBentryALTinterwordstretchfactor}{4}
\providecommand{\BIBentryALTinterwordspacing}{\spaceskip=\fontdimen2\font plus
\BIBentryALTinterwordstretchfactor\fontdimen3\font minus
  \fontdimen4\font\relax}
\providecommand{\BIBforeignlanguage}[2]{{%
\expandafter\ifx\csname l@#1\endcsname\relax
\typeout{** WARNING: IEEEtran.bst: No hyphenation pattern has been}%
\typeout{** loaded for the language `#1'. Using the pattern for}%
\typeout{** the default language instead.}%
\else
\language=\csname l@#1\endcsname
\fi
#2}}
\providecommand{\BIBdecl}{\relax}
\BIBdecl

\bibitem{Unity}
\BIBentryALTinterwordspacing
M.~{Sy}, C.~{Guo}, and J.~{Greco}, ``Unity game simulation: Find the perfect
  balance with {U}nity and {GCP},'' in \emph{Google for Games Developer
  Summit}, 2020. [Online]. Available:
  \url{https://events.withgoogle.com/gdc2020/}
\BIBentrySTDinterwordspacing

\bibitem{bergdahl}
J.~{Bergdahl}, C.~{Gordillo}, K.~{Tollmar}, and L.~{Gisslén}, ``Augmenting
  automated game testing with deep reinforcement learning,'' in \emph{2020 IEEE
  Conference on Games (CoG)}, 2020, pp. 600--603.

\bibitem{Agarwal20}
S.~Agarwal, C.~Herrmann, G.~Wallner, and F.~Beck, ``Visualizing ai playtesting
  data of 2d side-scrolling games,'' in \emph{Proceedings of IEEE Conference on
  Games}, aug 2020.

\bibitem{Stahlke20}
S.~Stahlke, A.~Nova, and P.~Mirza-Babaei, ``Artificial players in the design
  process: Developing an automated testing tool for game level and world
  design,'' in \emph{Proceedings of the Annual Symposium on Computer-Human
  Interaction in Play (CHI PLAY '20)}.\hskip 1em plus 0.5em minus 0.4em\relax
  New York, NY, USA: Association for Computing Machinery, 2020, p. 267–280.

\bibitem{Muratori18}
\BIBentryALTinterwordspacing
C.~Muratori, ``Killing the walk monster [{C}onference presentation],'' in
  \emph{BIC Festival}, 2018. [Online]. Available:
  \url{https://caseymuratori.com/blog_0032}
\BIBentrySTDinterwordspacing

\bibitem{ZhanAS19}
Z.~Zhan, B.~Aytemiz, and A.~M. Smith, ``Taking the scenic route: Automatic
  exploration for videogames,'' in \emph{KEG@AAAI}, ser. {CEUR} Workshop
  Proceedings, vol. 2313.\hskip 1em plus 0.5em minus 0.4em\relax CEUR-WS.org,
  2019, pp. 26--34.

\bibitem{Wuji19}
Y.~Zheng, X.~Xie, T.~Su, L.~Ma, J.~Hao, Z.~Meng, Y.~Liu, R.~Shen, Y.~Chen, and
  C.~Fan, ``Wuji: Automatic online combat game testing using evolutionary deep
  reinforcement learning,'' in \emph{Proceedings of the 34th IEEE/ACM
  International Conference on Automated Software Engineering}.\hskip 1em plus
  0.5em minus 0.4em\relax IEEE Press, 2019, p. 772–784.

\bibitem{ChangAS19}
K.~{Chang}, B.~{Aytemiz}, and A.~M. {Smith}, ``Reveal-more: Amplifying human
  effort in quality assurance testing using automated exploration,'' in
  \emph{2019 IEEE Conference on Games (CoG)}, 2019, pp. 1--8.

\bibitem{holmgaard2018automated}
C.~Holmg{\aa}rd, M.~C. Green, A.~Liapis, and J.~Togelius, ``Automated
  playtesting with procedural personas through mcts with evolved heuristics,''
  \emph{IEEE Transactions on Games}, vol.~11, no.~4, pp. 352--362, 2018.

\bibitem{Chang_Smith_2020}
K.~Chang and A.~Smith, ``Differentia: Visualizing incremental game design
  changes,'' \emph{Proceedings of the AAAI Conference on Artificial
  Intelligence and Interactive Digital Entertainment}, vol.~16, no.~1, pp.
  175--181, Oct. 2020.

\bibitem{schmidhuber1991possibility}
J.~Schmidhuber, ``A possibility for implementing curiosity and boredom in
  model-building neural controllers,'' in \emph{Proc. of the international
  conference on simulation of adaptive behavior: From animals to animats},
  1991, pp. 222--227.

\bibitem{pathak2017curiosity}
D.~{Pathak}, P.~{Agrawal}, A.~A. {Efros}, and T.~{Darrell}, ``Curiosity-driven
  exploration by self-supervised prediction,'' in \emph{2017 IEEE Conference on
  Computer Vision and Pattern Recognition Workshops (CVPRW)}, 2017.

\bibitem{Aubret19}
A.~Aubret, L.~Matignon, and S.~Hassas, ``A survey on intrinsic motivation in
  reinforcement learning,'' \emph{CoRR}, vol. abs/1908.06976, 2019.

\bibitem{schulman2017proximal}
J.~Schulman, F.~Wolski, P.~Dhariwal, A.~Radford, and O.~Klimov, ``Proximal
  policy optimization algorithms,'' \emph{arXiv preprint arXiv:1707.06347},
  2017.

\bibitem{Bellemare16}
M.~Bellemare, S.~Srinivasan, G.~Ostrovski, T.~Schaul, D.~Saxton, and R.~Munos,
  ``Unifying count-based exploration and intrinsic motivation,'' in
  \emph{Advances in Neural Information Processing Systems}, vol.~29.\hskip 1em
  plus 0.5em minus 0.4em\relax Curran Associates, Inc., 2016, pp. 1471--1479.

\bibitem{LaValle1998}
S.~LaValle, ``Rapidly-exploring random trees : a new tool for path planning,''
  \emph{Technical Report TR 98-11, Computer Science Department, Iowa State
  University}, 1998.

\end{thebibliography}
\bibliographystyle{IEEEtran}

\end{document}